\documentclass[11pt]{article}
\usepackage{gal-basic}

\title{Decision-Making under Miscalibration}

\author{Guy N.\ Rothblum
\\Weizmann Institute of Science\\\texttt{rothblum@alum.mit.edu}
\and
Gal Yona
\\Weizmann Institute of Science\\\texttt{gal.yona@weizmann.ac.il}}

\begin{document}

\maketitle
\begin{abstract} 
ML-based predictions are used to inform consequential decisions about individuals. How should we use predictions  (e.g., risk of heart attack) to inform downstream binary classification decisions (e.g., undergoing a medical procedure)? When the risk estimates are perfectly calibrated, the answer is well understood: a classification problem’s cost structure induces an optimal treatment threshold $\js$. In practice, however, some amount of miscalibration is unavoidable, raising a fundamental question: how should one use potentially miscalibrated predictions to inform binary decisions? 

We formalize a natural (distribution-free) solution concept: given anticipated miscalibration of $\alpha$, we propose using the threshold $j$ that minimizes the worst-case regret over all $\alpha$-miscalibrated predictors, where the regret is the difference in clinical utility between using the threshold in question and using the optimal threshold in hindsight. We provide closed form expressions for $j$ when miscalibration is measured using both expected and maximum calibration error, which reveal that it indeed differs from $\js$ (the optimal threshold under perfect calibration). We validate our theoretical findings on real data, demonstrating that there are natural cases in which making decisions using $j$ improves the clinical utility.
\end{abstract}

\section{Introduction}

Powered by advances in machine learning, ML-based predictions are used to inform increasingly consequential decisions about individuals. We consider a  setup in which a single risk predictor (e.g., estimating the risk of a cardiovascular event in a 10-year followup period) is used to inform multiple downstream binary classification decisions  (e.g., whether to prescribe a certain medication, or have the individual undergo a medical procedure). Naturally, these problems may differ in their \emph{cost structure} (the relative costs of correct and incorrect predictions). For example, a false positive can be quite costly in the context of a potentially dangerous medical procedure, but less costly in the context of prescribing a medication without significant side effects\footnote{For the remainder of this paper we have this example in mind and therefore refer to the binary decisions as treat/don't treat decisions.}. The cost structure determines the \emph{utility} of a proposed classifier; specifically, we consider the Net Benefit  \cite{vickers2006decision}, defined as a weighted combination of the fraction of true positive and false positive predictions, with the weights determined by the cost structure.

With this setup in mind, how should predictions be translated into binary treatment decisions? When the risk predictor is \emph{calibrated}, this question is well understood: we should treat individuals whose predicted risk exceeds the {\em therapeutic threshold}\footnote{The therapeutic threshold is the smallest {\em true} risk level (true probability of a positive outcome) where treatment gives non-negative utility. If a predictor is calibrated, then applying the therapeutic threshold to the {\em predicted} risks maximizes the utility. The well-known fact that thresholding the predictions at 0.5 minimizes the $\ell_1$ (i.e., 0/1) loss (symmetric costs) is a special case.}, $\js$ (a function of the cost structure of the classification problem) \cite{pauker1975therapeutic}. Here, (perfect) calibration is the requirement that for each prediction ``category'' $v \in [0,1]$, of the individuals for whom the predicted risk was $v$, exactly a $v$-fraction actually receive positive outcomes. Previous work demonstrated, both in real-world scenarios and through simulations, that applying $\js$ to \emph{miscalibrated} risk predictions  negatively effects the clinical utility of downstream decisions \cite{collins2012predicting, van2015calibration, otlecs2022development}; see Section \ref{sec:related} for an extended discussion.

\subsection{From predictions to decisions under miscalibration}

While the above discussion stresses the importance of calibration as a requirement for risk models, in practice 
\emph{some amount of miscalibration is unavoidable} (e.g., as a result of generalization errors or distribution shifts). The inevitability of some degree of calibration errors, and the potential for negative downstream impact, necessitate re-evaluating how risk predictions are translated into decisions in the presence of miscalibration. 


Our starting point is the following question: if $\js$ is the optimal threshold to apply to a predictor $p$ that is \emph{perfectly} calibrated (irrespective of both the target distribution and the specifics of the predictor $p$), is there an analogously optimal threshold $\hat{j}$  to apply to some predictor $p'$, when all we are willing to assume about $p'$ is that it is at most $\alpha$-miscalibrated w.r.t the (unknown) target distribution over individuals and outcomes?   This raises many natural questions: which solution concept should we use to determine optimality in this context? What should this threshold be? And when is it actually different from $\js$?

To build intuition, consider a classification task for which the 
 therapeutic threshold is $\js=0.05$ (specifically, this means the benefit from a true positive prediction is 19 times larger than the harm from a false positive prediction; see the discussion in Section \ref{sec:prelims:utility}). Should we treat  individuals whose {\em predicted} risk is 0.04? Under perfect calibration we would be guaranteed that these individuals would not benefit from treatment; but what if we anticipate some amount of miscalibration?  If we use $\js$, but our predictor is under-estimating the risk (the probability of  positive outcomes w.r.t the target distribution), we will regret not treating. Perhaps we should use a lower threshold, e.g. $\hat{j} = 0.03$? But if we use $\hat{j}$ and our predictor is {\em over}-estimating the risk, we will regret treating! Balancing these competing scenarios is tricky, since the non-symmetric errors within a certain miscalibration bound $\alpha$ (over- or under-estimation) may have different consequences in our context. Given these complex considerations, what treatment threshold should we use?
 
Our contributions are:

\textbf{(1) A new solution concept: minimizing the worst-case regret under miscalibration.} We formalize the above intuition by defining the \emph{regret} from thresholding a predictor $p$ at a threshold $j \in [0,1]$ rather than at a threshold $j' \in [0,1]$, when the distribution on individuals-outcome pairs is given by $D$
. The regret is the difference between the expected utility obtained by using $j'$ and the expected utility obtained by using $j$ (where the expectation is over the distribution $D$).
Since the only thing we are assuming about the predictor in question is that it is at most $\alpha$ miscalibrated w.r.t the (unknown) target distribution,  we propose that decisions should be made using the treatment threshold $j$ that minimizes the {\em worst-case regret}, defined as the maximal regret that can be experienced, over all the possible alternative thresholds $j'$, and over all possible predictors $p$ and distributions $D$ (that satisfy that $p$ is at most $\alpha$-miscalibrated on $D$). 

We emphasize two points: First, the regret-minimizing threshold is, by design (and similarly to the therapeutic threshold $\js$), a {\em distribution free} notion: it does not depend on either the distribution $D$ or the predictor in question. Rather, it is only a function of the anticipated error rate $\alpha$ and of the cost structure (as manifested in the treatment threshold $\js$). Second,  looking ahead, the therapeutic threshold $\js$ does not always minimize the regret: it diverges from the regret-minimizing threshold when we anticipate errors ($\alpha > 0$) and when the problem's cost structure is sufficiently asymmetrical (see below).  

Our notion of regret is related to the notion of {\em Clinical Harm}: a performance measure from the decision-analytic literature \cite{van2016calibration}. The clinical harm of a classification model is the difference between its utility and the utility of (the better of) two naive baselines: treating all individuals (treatment threshold 0) and treating none (threshold 1). We measure regret with respect to {\em all} possible alternative threshold (not just 0 and 1) and thus the regret for using a threshold is always at least as large as its clinical harm. Bounding the worst-case regret also implies bounds on the clinical harm. 

\textbf{(2) Simple closed-form expressions for the regret-minimizing thresholds.}
We provide simple (and efficiently-computable) closed-form solutions for computing the regret-minimizing threshold, when the miscalibration is quantified using the standard measures of maximum and expected calibration errors (Definitions \ref{def:mce} and \ref{def:ece}). This is our main technical contribution. We note that the regret-minimizing threshold is the solution to a min-max problem over an exponential space of possibilities (in particular, all possible predictors and distributions that conform to the miscalibration bound), so it is infeasible to compute it by enumerating over all possibilities.

Our derivation highlights that the optimal threshold naturally follows a ``conservatism in the face of uncertainty'' approach. Recall the example above, where $\js=0.05$ and the predicted risk is 0.04: intuitively, it made sense to hedge our bets and treat these individuals, since the potential benefit if their risk was underestimated is much larger than the potential harm if the risk was over-estimated. The regret-minimizing threshold gives a rigorous explanation and formalization for this intuition.

On a technical level, we obtain the closed-form expressions in two steps. First, we show that the regret can always be maximized by a predictor that comes from a very restricted class, where all the individuals have one of (at most) two possible risk levels. We then derive a closed-form expression for computing a regret-minimizing threshold over this restricted class. See Section \ref{sec:general} for further details.

\textbf{(3) Experimental validation.} We validate our findings on the task of 
risk prediction for adverse events due to traumatic brain injury. We demonstrate natural scenarios in which using the more conservative (regret-minimizing) threshold indeed improves the expected clinical utility of the predictor in question, see Section \ref{sec:validation}. 
 
\textbf{Fairness implications.} 
Our work also has implications for recent developments in the area of \emph{algorithmic fairness}. In this work we considered calibration on the population as a whole;
Starting with \cite{hkrr}, a series of recent works aim to obtain {\em multi-calibrated} risk predictors, whose calibration errors are bounded for a rich collections of protected demographic subgroups. Our work is motivated by (and is closely related to) multi-calibration in two respects. First, the connection between calibration and decision-analytic notions such as Clinical Harm crystallizes a concrete subgroup harm that multi-calibrated predictions prevent: specifically, the existence of a protected subgroup for which the utility obtained by \emph{using} the predictor is lower than the utility that could have been obtained \emph{without using} the predictor. 
Second, our work provides a principled methodology for translating multi-calibrated predictions into decisions, minimizing the worst-case regret experienced by each of the subgroups.

\textbf{Organization.} The rest of the manuscript is organized as follows. In Section \ref{sec:prelims} we define our setup and the notions of calibration and utility we will use. In Section \ref{sec:harm-to-regret} we define our notion of regret. In Section \ref{sec:perfect} we show that under perfect calibration, treating at $\js$ guarantees non-negative regret, but that under miscalibration, the regret can be negative (and significant, depending on the cost structure). We study regret-minimizing thresholds under miscalibration in Section \ref{sec:general}. We validate our findings on real-data in Section \ref{sec:validation} and conclude with additional related work (Section \ref{sec:related}) and natural directions for future exploration (Section \ref{sec:discuss}). Full proofs are deferred to the appendix.

\section{Preliminaries}
\label{sec:prelims}

Let $\X$ denote the space of  covariates and $Y\in\{0,1\}$ the modeled event (e.g. heart attack within 10 years). Let $\D$ denote an unknown distribution on $\X$. 
We distinguish between \emph{risk predictors} (whose goal is to predict the probability of the target event given covariates) and \emph{classifiers} (denoting the binary decisions that are informed by the risk estimate). 

\textbf{Risk predictors.} Throughout, we assume the $[0,1]$ risk prediction interval is discretized to a grid of width $1/m$, $G_{m}=\{\frac{1}{m},\frac{2}{m},\dots,\frac{m-1}{m}\}$, where $m\in\N$ is an external parameter controlling the fine-ness of the grid. Hence, predictors are a mapping $p: \X \to G_m$. As a convention, we  use $p^\star$ to denote predictors that determine the data-generation process (specifying the probability of positive outcome given covariates) and $p$ to denote estimates. For a predictor $p$,  $\mu_{p,i}$ denotes the probability mass of individuals whose prediction is $i/m$: $\mu_{p,i} = \Pr_{x \sim \D}[p(x) = i/m]$. In Section \ref{sec:general} we  use $\Xi$ to denote all the set of all ``legal'' level-sets, i.e. $\set{\mu_{p,i}}_{i=1}^m$ such that each $\mu_{p,i}$ is non-negative and all the $\mu_{p,i}$'s sum to 1.

\textbf{Classifiers.}  A classifier is a mapping $h: \X \to \set{0,1}$. Given that the outcomes are generated according to $p^\star$, we use $\mathtt{TP}_{\ps}(h)$ and $\mathtt{FP}_{\ps}(h)$ to denote the fraction of \emph{true positive} and \emph{false positive} classifications in the target population\footnote{Note that these are different from the true positive and false positive \emph{rates} which measure the conditional probability, e.g. $\Pr_{x\sim \D, y\sim \ps(x)}[h(x) = 1 \vert y = 0]$.}: $ \mathtt{TP}_{\ps}(h) = \Pr_{x\sim \D, y\sim \ps(x)}[h(x) = 1 \land y = 1]$ and $\mathtt{FP}_{\ps}(h) = \Pr_{x\sim \D, y\sim \ps(x)}[h(x) = 1 \land y = 0]$.


In this work, our focus is on classifiers derived from the risk predictions. I.e., in which $h(x) \in \set{0,1}$ is some function of $p(x)$ (and possibly $x$ itself). We restrict our attention to family of transformations that determine treatment by thresholding the risk predictor at a \emph{fixed threshold}; we use $h_{p,j}$ to denote the binary decisions obtained by thresholding $p$ at threshold $j/m$: $h_{p,j}(x) :=\textbf{1}[p(x)>j/m]$.

In this section we review two different ways to quantify the performance of risk predictors. \emph{Clinical harm} (Section \ref{sec:prelims:utility}) explicitly takes into account the downstream decisions informed by the predictions via a simplified utility-based analysis, and measures the potential decrease in performance relative to two naive benchmarks. \emph{Calibration} (Section \ref{sec:prelims:calibration}), on the other hand, does not take into account the downstream decisions and instead asks that the risk predictor outputs scores that can meaningfully be interpreted as probabilities.

\subsection{Calibration}
\label{sec:prelims:calibration}

Perfect calibration requires that for every value $v\in [0,1]$, the true expectation of the outcomes among those who receive prediction $v$ is exactly $v$. In practice, as mentioned, we don't expect predictors to be perfectly calibrated. We now discuss two different ways to measure \emph{approximate} calibration. We use notions from the literature but our presentation will be slightly different: Usually, the data generating process (for us, denoted by $\ps$) is considered as fixed and so calibration is only a property of the predictor in question. Looking ahead, we will instead think of calibration as a joint property of both $p$ and $\ps$ (formally, as a \emph{relation} over pairs of predictors). In our setup, $p$ is perfectly calibrated w.r.t $\ps$ if for every $i \in [m]$ for which $\mu_{p,i} > 0$,
\begin{equation*}
    \underbrace{\Pr_{x\sim D, y\sim \ps(x)}[y \vert p(x) = i/m]}_{\triangleq\,\,  \ytilde} = i/m
\end{equation*}

The calibration error on the $i$-th bin is therefore $\card{\ytilde - i/m}$. \emph{Expected calibration error} bounds the weighted sum of calibration errors, where the $i-$th bin is weighted according to the fractional mass of $\D$ that ``lands'' in this bin according to the predictor in question. \emph{Maximum calibration error} instead bounds each $\card{\ytilde - i}$ separately. 

\begin{definition}[Expected Calibration Error (ECE) \cite{naeini2015obtaining}]
\label{def:ece}
Fix $\gamma > 0$. We say that a predictor $p$ has an expected calibration error of $\gamma$ w.r.t $\ps$  if $\sum_{i=1}^m \mu_{p,i}\cdot \card{\ytilde - i} \leq \gamma $.
\end{definition}

\begin{definition}[Maximum Calibration Error (MCE) \cite{naeini2015obtaining}]
\label{def:mce}
Fix $\gamma > 0$. We say that a predictor $p$ has a maximum calibration error of $\gamma$ w.r.t $\ps$  if for every $i \in [m]$ for which $\mu_{p,i} > 0$, we have $\card{\ytilde - i} \leq \gamma $.
\end{definition}

We note that both ECE and MCE depend on the underlying distribution via the level sets $\set{\mu_{p,i}}_{i=1}^m$. We thus use $(p,\ps) \in R_{ECE}(\gamma)$ and $(p,\ps) \in R_{MCE}(\gamma)$  as shorthand notation for the ECE and MCE relations, when the relevant $\set{\mu_{p,i}}_{i=1}^m$'s are clear from context.

We also note that MCE is a stronger requirement than ECE:  keeping $\set{\mu_{p,i}}_{i=1}^m$ fixed, $(p,\ps) \in R_{MCE}(\gamma)$ implies $(p,\ps) \in R_{ECE}(\gamma)$.

\subsection{Utility-based metrics}
\label{sec:prelims:utility}

In this section we give a self-contained introduction to decision-curve analysis. We begin with a high-level presentation of the ideas and delay formal definitions to Section \ref{sec:generalized_net_benefit}, where we discuss several extensions that will be important to our work.

\textbf{Background.} Predictive models are often evaluated using notions of accuracy, such as calibration and various measures of discrimination (sensitivity, specificity, AUC). However, these methods often have little clinical relevance: e.g., how high an AUC is high enough to justify clinical use of a prediction model?  \cite{baker2009using} Naturally, the answer depends on the consequences
of the particular clinical decisions informed by the predictive model. Decision analytic methods explicitly consider the clinical consequences of decisions.  Let $U_{TP}$, $U_{FP}$,  $U_{TN}$ and  $U_{FN}$ denote the utilities for all combinations of treatment decisions and disease outcomes\footnote{For a concrete example, \cite{gail2005criteria} give the following numeric example for colorectal cancer: $U_{FN} = -100$, for the possibility of death and morbidity due to failing to detect colorectal cancer; $U_{FP}=-1$, for the risk of bleeding or perforation of the colon; $U_{TP}=-11$, for the risk of bleeding or perforation of the colon and the lowered chance of death or morbidity from colorectal cancer due to early detection; and finally, $U_{TN}$ is set to zero as a reference value.}.

Given a predictor $p$ and costs $\mathcal{U} = \set{U_{TP}, U_{FP}, U_{FN}, U_{TN}}$, how should treatment decisions be made? Consider an individual that receives a prediction of $i/m$. Recalling the short-hand notation $\ytilde \triangleq \Pr_{x\sim \D, y\sim \ps(x)}[y=1 \vert p(x) = v]$, their expected utility from opting to receive treatment is $\ytilde\cdot U_{TP} + (1-\ytilde)\cdot U_{FP}$; similarly, 
their expected utility from opting to not receive  treatment is $\ytilde\cdot U_{FN} + (1-\ytilde)\cdot U_{TN}$. Thus, from a utility theory perspective, the optimal threshold, which we denote $\js$ and refer to as the \emph{therapeutic threshold}, is the point for which these two utilities are equal: $   \ytildejs \cdot U_{TP} + (1-\ytildejs)\cdot U_{FP} = \ytildejs \cdot U_{FN} + (1-\ytildejs)\cdot U_{TN}$. Under \emph{perfect calibration} $\ytildejs = \js/m$; plugging this in and re-organizing, we obtain 
 $\frac{m-\js}{\js} = \frac{U_{TP} - U_{FN}}{U_{TN} - U_{FP}}$. In other words, under perfect calibration, we have the following relationship between the therapeutic threshold $\js$ and the derived costs:
\begin{equation}
\label{eqn:j_p_l}
    \frac{m-\js}{\js} = \frac{P}{L}
\end{equation}

where the \emph{profit} $P \triangleq U_{TP} - U_{FN}$ is the difference in utilities from making a positive instead of a negative prediction for disease among those \emph{with} the disease, and the \emph{loss} $L \triangleq U_{TN} - U_{FP}$ is the negative of the difference in utilities from making a positive instead of a negative prediction for disease among those \emph{without} the disease.

\textbf{Net Benefit, Clinical Utility and Clinical Harm.} 
Decision curve analysis \cite{vickers2006decision}  is a simplified utility-based analysis: instead of specifying the full costs $\mathcal{U} = \set{U_{TP}, U_{FP}, U_{FN}, U_{TN}}$, it only relies on the relative values $P$ and $L$\footnote{For example, in the example of colorectal cancer discussed above \cite{gail2005criteria}, the profit is $P=89$ the loss is $L=1$.}. By fixing the profit at $P=1$, the relative harm of a False Positive prediction, following Equation (\ref{eqn:j_p_l}), is given by $L = -\frac{\js}{m-\js}$.  \cite{vickers2006decision} proceed to define the \emph{Net Benefit} of the predictor $p$ as the weighted average of False Positives and True Positives of binary classifier $h_{p, \js}$ (obtained by thresholding the risk scores at $\js$): $\mathtt{NetBenefit}(p) = P \cdot \mathtt{TP}_{\ps}(h_{p, \js})- L \cdot \mathtt{FP}_{\ps}(h_{p, \js})=    \mathtt{TP}_{\ps}(h_{p, \js})-\frac{\js}{m-\js} \cdot \mathtt{FP}_{\ps}(h_{p, \js})$.

This is the called Net Benefit of the model as it reflects the ``effective'' fraction of True Positives: the fraction of True Positive predictions, minus the number of False Positive predictions -- as ``valued'' in terms of True Positives.

Suppose we calculated the Net Benefit of a risk model $p$ at some threshold to be, say, 0.151. How ``good'' is this value? Is it high enough to justify treatment using the model? One simple ``sanity check'' is that the Net Benefit of the model is better than the Net Benefit obtained by two naive alternatives, that could have been carried out \emph{without} the risk prediction model: (i) to treat everyone, and (ii) to treat no one. The \emph{Clinical Utility}  of a prediction model $p$ is the excess Net Benefit over the better of these two simple baselines, and a model is said to be \emph{Clinically harmful} if there exists a treatment threshold for which the clinical utility is negative \cite{van2016calibration}. When the risk threshold is varied, the Net Benefit
of the model, the Net Benefit of the default strategies and the model's Clinical Utility can be plotted
by risk threshold to obtain a \emph{decision curve}; see Section \ref{sec:validation} for examples and \cite{kerr2016assessing, vickers2019simple} for an overview of reading such decision curves from a practitioner's viewpoint. 

\subsection{Symmetric and Disentangled Net Benefit}
\label{sec:generalized_net_benefit}
In this section we describe two extensions of the notion of the Net Benefit that will be critical to our study of the effects of miscalibration on the Net Benefit: the first defines a symmetric version of the Net benefit, and the second 
generalizes the Net Benefit to classifiers derived by thresholds other than $\js$. We conclude the section with a derivation of a closed-form expression for the generalized Net Benefit, which we will use extensively in Section \ref{sec:general}.

\textbf{Symmetric Net Benefit.} A technical issue that will be problematic for us moving forward is the fact that the Net Benefit as originally defined in \cite{vickers2006decision} is not symmetric: it takes values in the range $(-\infty, \beta)$, where $\beta$ is the base rate in the target population. This is an artifact of the (arbitrary) choice to fix the profit (rather than the loss) at $1$. To address this, we will consider the Net Benefit w.r.t new \emph{symmetric} relative costs $P'$ and $L'$, ensuring that the ratio remains unchanged, $P/L = P'/L'$ (recall, from Equation \ref{eqn:j_p_l}, only the ratio of the costs matters in this analysis). Specifically, we will ensure that the \emph{lower} of the two costs is set to 1:

\begin{itemize}
    \item If $\js < m/2$, then $P > L$, so we set $L'=1$ and obtain $P' = \frac{m-\js}{\js}$.
    \item If $\js > m/2$, then $P < L$, so we set $P'=1$ and obtain $L' = \frac{j*}{m-j*}$.
\end{itemize}

\textbf{Disentangled Net Benefit. } It's crucial to note that in the above discussion, the therapeutic threshold $\js$ served \textbf{two distinct roles}:  on one hand, it defined the transformation from predictions ($p$) to binary decisions ($h_{p, \js}$); on the other hand, it also implicitly captured the costs in question, and thus determined how the derived binary classifier $h_{p, \js}$ was evaluated. In other words, the existing literature on Net Benefit implicitly assumes that the transformation from predictions to decisions will use the therapeutic threshold, $\js$. As discussed, an important component of our work is considering making binary decisions using thresholds \emph{other than $\js$}. This is a natural perspective when miscalibration is involved, since the derivation of $\js$ as the ``optimal'' threshold was under the assumption of perfect calibration. It is therefore crucial for us to disentangle  these two roles of $\js$. We will work with the following more general notion of Net Benefit, that quantifies the Net Benefit of making predictions using a threshold $j$ when the \emph{actual costs} are implied by $\js$:

\begin{definition}[(Disentangled) Net Benefit]
Fix a predictor $p$. The Net Benefit of making predictions using a threshold $j$ when the therapeutic threshold is $\js$ is 
\begin{equation}
\label{eqn:net_benefit_P_L}
    \Lambda(p, j; \ps, \js) =   P' \cdot \mathtt{TP}_{\ps}(h_{p, j})- L' \cdot \mathtt{FP}_{\ps}(h_{p,j}) 
\end{equation}

\end{definition}

We can combine Equation \ref{eqn:net_benefit_P_L} with the definitions of the symmetric costs $P'$ and $L'$ from the previous section to obtain a useful expression for the disentangled Net Benefit:

\begin{lemma}
\label{lemma:net_benefit}
For every $\set{\mu_{p,i}}_{i=1}^m$, $p, \ps$ and $j, \js$,

\begin{equation}
    \Lambda(p,j; \ps, \js)=\sum_{i>j}\mu_{p,i}\cdot\frac{m\cdot \tilde{y}_{p,\ps,i}-\js}{\min\left\{ m-\js,\,\,\js\right\} }
\end{equation}
\end{lemma}

See Appendix \ref{appendix:nb_lemma_proof} for the proof.

\section{From Clinical Harm to General Regret}
\label{sec:harm-to-regret}

We now discuss a final extension to decision-curve analysis, in which we apply  ideas from online learning and regret minimization to extend the definition of Clinical Harm. 
Recall that the Clinical Harm from making decisions using a predictor $p$ and threshold $j$ is the Net Benefit of the better of the two naive strategies (treat all, and treat none) minus the Net Benefit of the model. Since \emph{treat all} is like thresholding $p$ at 0 and and \emph{treat none} is like thresholding $p$ at 1, we can write the clinical harm succinctly as $\max \set{ \Lambda(p,0; \ps, \js),  \,\, \Lambda(p,1; \ps, \js)} - \Lambda(p,j; \ps, \js)$.

With this in mind, it is natural to ``measure'' $j$ not only against the two specific thresholds $\set{0,1}$, but also against \emph{every} constant threshold. We refer to this as the \emph{regret} incurred by making decisions using a predictor $p$ and threshold $j$, akin to the notion of regret from online learning (where the ``benchmark'' class we compete against is the class of all constant thresholds):

\begin{equation}
    \label{eqn:regret}
    \mathtt{regret}(p,j; \ps, \js) = 
    \max_{j'} \Lambda(p,j'; \ps, \js) -  \Lambda(p,j; \ps, \js)
\end{equation}

In Section \ref{sec:general} we study the question of minimizing the worst-case regret under miscalibration.

\section{Treating at the therapeutic threshold}
\label{sec:perfect}

In this section, we formally study the guarantees obtained by thresholding $p$ at the therapeutic threshold $\js$. In particular, we show that: (i) when $p$ is \emph{perfectly calibrated}, using $\js$ guarantees \emph{no regret} (in that the regret from Equation \ref{eqn:regret} is non-positive); (ii) when $p$ is $\alpha$-miscalibrated, using $\js$ can lead to regret that scales like $\alpha \cdot m$.

\begin{lemma}[No regret under perfect calibration.]
\label{lemma:perfect}
\begin{equation*}
    (p,\ps) \in R_{ECE}(0) \Longrightarrow \mathtt{regret}(p,\js; \ps, \js) \leq 0
\end{equation*}
\end{lemma}

See Appendix \ref{appendix:lemma_proof} for the proof.

\textbf{Example} (Regret under $\alpha$-miscalibration).
Consider the following two  constant predictors: $\ps \equiv 1-\alpha$ and $p \equiv 1$. Clearly, $(p,\ps) \in R_{ECE}(\alpha)$. Consider the Clinical Utility of treating at the therapeutic threshold when the latter is equal to $\js = m-1$. Note that since $p=1$, the Net Benefit of the model is essentially the Net Benefit of the ``treat all'' strategy, which is $ ((1-\alpha) \cdot 1) - (\alpha \cdot \frac{\js}{m-\js}) = 1 - \alpha \cdot m$. On the other hand, the Net Benefit of the ``treat none'' strategy is 0. Hence the Clinical Utility of the model is $1-\alpha\cdot m$, and the regret of using $p$ with threshold $\js$ is lower bounded by $1-\alpha \cdot m$.

\section{Decision-making under miscalibration}
\label{sec:general}

Intuitively, the above result shows that under miscalibration, there are extreme cases in which using the therapeutic threshold $\js$ ``as is'' could be very costly. Our example above highlights two natural ways to reduce this ``worst case'' clinical harm: the first is to obtain predictors with stronger calibration guarantees to begin with, and the second is to use coarser predictions (effectively limiting the largest expressible cost of a misclassification in this framework). In practice, some amount of miscalibration is unavoidable, and having the ability to make granular predictions is important. We therefore turn our attention to a third ``knob'': the mechanism by which we translate predictions to decisions. Specifically, we ask: 
\vspace{-0.5em}
\begin{center}
    \emph{Given $\alpha > 0$ and the therepeutic threhsold $\js$, which threhsold minimizes the worst-case regret under $\alpha$-miscalibration? }
\end{center}

Giving rise to the following min-max optimization problem:

\begin{equation}
\label{eqn:optimization_problem}
    \hat{j}(\alpha,m,\js) \in \arg\min_{j}\underbrace{\max_{\substack{\set{\mu_{p,i}}_{i=1}^m \in \Xi \\ (p,p^{\star})\in R(\alpha)}} \mathtt{regret}(p,j; \ps, \js)}_{\triangleq c(j; \js, R(\alpha))}
\end{equation}

Here, $R(\alpha)$ is a miscalibration relation (e.g. ECE or MCE) and we define the ``cost'' of a threshold, $c(j;\js, R(\alpha))$, as the maximal regret that we might experience by using it. The maximum is effectively over any distribution $D$ on the domain $\X$ and every predictors $p$ and $\ps$ that are within the allowed miscalibration level; in practice, since the only effect the distribution $D$ has on this expression is via the level sets of the predictor $p$, we replace $D$ with the masses of the level sets, $\set{\mu_{p,i}}_{i=1}^m$. Recall that $\Xi$ ensures these choices are legal (non-negative and sum to 1).

We refer to a solution of Equation (\ref{eqn:optimization_problem}) as the \emph{conservative therapeutic threshold}.We note that the fact we have shown that using $\js$ can cause $\alpha \cdot m$ clinical harm does not yet imply that $\js$ itself is \emph{not} a solution to (\ref{eqn:optimization_problem}); in principle, we anticipate that in this setup every choice of threshold $j$ would give rise to some regret, and so using $\js$ itself still could be the best threshold, in a \emph{regret minimization} sense.

\begin{figure}[tbp]
    \centering
    \includegraphics[width=0.6\textwidth]{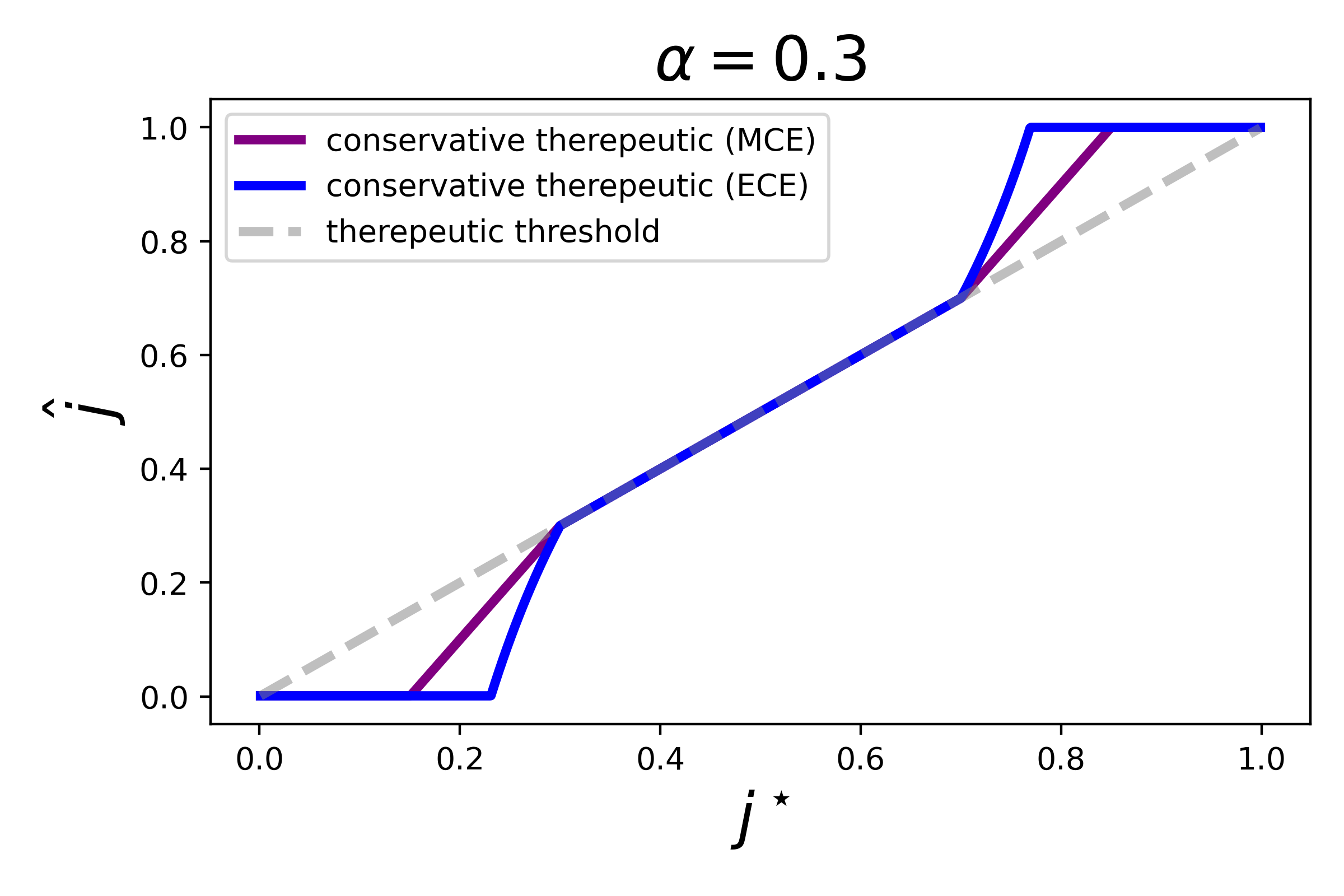}
    \caption{Plotting the conservative therapeutic thresholds $\hat{j}_{MCE}$ and $\hat{j}_{ECE}$ (see Theorem \ref{thm:optimal})  vs the standard therapeutic threshold.}
    \label{fig:j_hat}
\end{figure}

Our main technical contribution is a closed-form expression for the optimal threshold (a solution to Equation (\ref{eqn:optimization_problem})), when miscalibration is measured using either maximum calibration error:

\begin{theorem}[Optimal thresholds under miscalibration]
\label{thm:optimal}
Fix $m\in \N$ and $\alpha > 0$. Then $\hat{j}_{MCE}(\alpha,m,\js)$ (the optimal threshold when the MCE can be as large as $\alpha$) is given by

\begin{equation*}
\begin{cases}
    0 & 0\leq \js\leq \frac{\alpha}{2}\cdot m\\
    2\js-\alpha m & \frac{\alpha}{2}\cdot m \leq \js\leq\alpha m\\
    \js & \alpha m\leq \js\leq(1-\alpha)m\\
    2\js-(1-\alpha)m & (1-\alpha)m\leq \js\leq(1-\frac{\alpha}{2})\cdot m\\
    m & (1-\frac{\alpha}{2})\cdot m\leq \js\leq m
    \end{cases}   
\end{equation*}

and $\hat{j}_{ECE}(\alpha,m,\js)$ (the optimal threshold when the ECE can be as large as $\alpha$) is given by

\begin{equation*}
        \begin{cases}
    \max\left\{ 1,\,\,(1+\alpha)\cdot m-\frac{\alpha m^{2}}{\js}\right\}  & \js<\alpha m\\
    \js & \alpha m<\js<(1-\alpha)m\\
    \min\left\{ m,\,\,\alpha m\cdot\frac{\js}{m-\js}\right\}  & \js>(1-\alpha)m
    \end{cases}
\end{equation*}

\end{theorem}
Theorem 1 reveals that irrespective of how we measure miscalibration and for every $\alpha > 0$,  $\js$ is indeed \emph{not} the regret-minimizing threshold (at least not everywhere). Intuitively, the expression for $\hat{j}$ reveals a natural ``conservatism in the face of uncertainty'' behaviour: the conservative threshold ``clips'' the decisions in an $\alpha$-region around the edges of the prediction interval, 
where the relative costs of false positive or false negative predictions are highest, and $\alpha$-miscalibration could be most detrimental. Figure \ref{fig:j_hat} demonstrates visually how
the optimal thresholds differ from $\js$, and also how they differ from one another. Intuitively, we see that $\hat{j}_{ECE}$ is even more ``conservative'' than $\hat{j}_{MCE}$. Intuitively, this makes sense -- bounding the expected calibration error of a predictor is a weaker guarantee than bounding the maximum calibration error.

See Appendix \ref{appendix:thm_proof} for the proof; we sketch the main idea below. For both notions, the proof consists of three parts:
\begin{enumerate}
    \item Reducing the original optimization problem (Equation \ref{eqn:optimization_problem}) into a simplified one, where $p$ is additionally constrained to be ``simple''. Formally, simple depends on how we measure miscalibration: for MCE, this is a constant predictor (i.e., that is supported on only a single value); and for ECE, this is an ``almost'' constant predictor (i.e., supported on at most two values).
    \item Showing that under the constraint that $p$ is ``simple'', we can carefully derive a closed-form expression for the maximal regret under miscalibration (cost).
    \item Showing that the minimizer of the above cost follows the expressions defined in the theorem statement.
\end{enumerate}

\begin{figure}[tbp]
    \centering
    \includegraphics[width=0.85\linewidth]{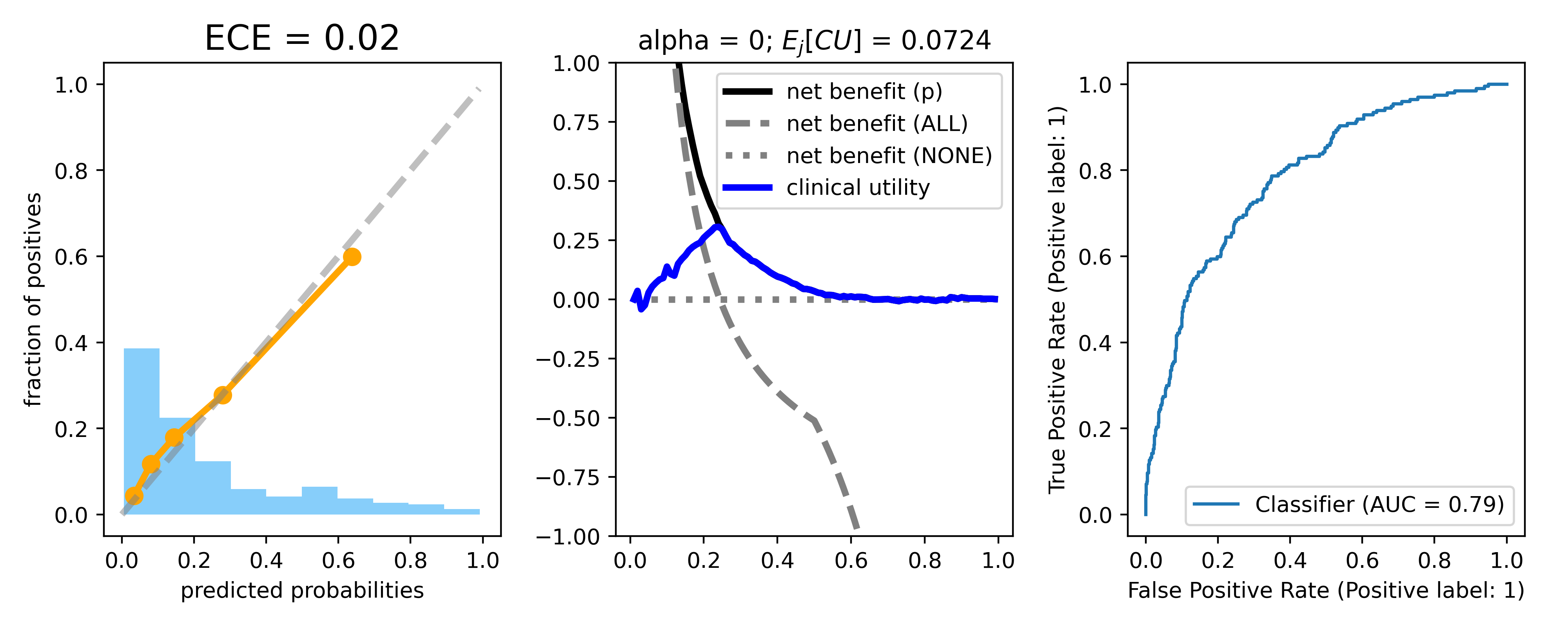}
    \caption{The baseline model is well-calibrated (left) and has an AUC of 0.79 (right). It has clinical utility in the range $[0.1, 0.4]$, as evident in the decision curves (middle). It is not clinically harmful, as for every possible threshold, the Net Benefit of the model exceeds the naive baselines, plotted in gray.}
    \label{fig:baseline}
\end{figure}

\section{Empirical validation}
\label{sec:validation}

We use\footnote{Our code is at \texttt{https://github.com/galyona/decision-making-under-miscalibration}} the Tirilazad trial data\footnote{Source: \texttt{{http://clinicalpredictionmodels.org/}}} for outcomes of traumatic brain injury ($n=2159$). The data consists of two trials, International ($n=1118$) and US ($n=1041$). The primary  outcome is the 6 months Glasgow Outcome Scale (ranges from 1 for dead, to 5 for good recovery), from which we derive a binary target of mortality at 6 months (23\% of the cases).  
We use logistic regression to fit a model to a random 80\% subset of the data. The baseline model obtains an AUC of ~0.8, and is overall well-calibrated. Figure \ref{fig:baseline}  demonstrates the calibration curves, decision curves and ROC curves, computed on the remaining 20\% of the data. The decision curves are computed w.r.t the therapeutic threshold, demonstrating that the model is clinically useful in the range $[0.1, 0.4]$.

Recall that in Section \ref{sec:general}, we derived the optimal thresholds when miscalibration is measured using \emph{maximum} and \emph{expected} calibration error (Theorem \ref{thm:optimal}) -- denoted  $j_{MCE}$ and $j_{ECE}$, respectively. Which threshold should we use? 
Empirically, we observe that for a fixed task, naively using $j_{MCE}$ may be over-conservative; We describe the details of this evaluation below. For the remaining experiments we therefore use $j_{ECE}$. We consider two natural setups in which miscalibration may occur: \emph{learning from small data}, in which the empirical calibration error (computed on the training set) may poorly reflect the calibration error on the underlying distribution; and a \emph{distribution shift} setting, in which we evaluate the model on subgroups that were not present in training time. We show that in both cases, there is naturally-occurring test time miscalibration, for which using the conservative therapeutic threshold can effectively ``guard against''.

 \begin{figure}[tbp]
    \centering
    \includegraphics[width=0.5\linewidth]{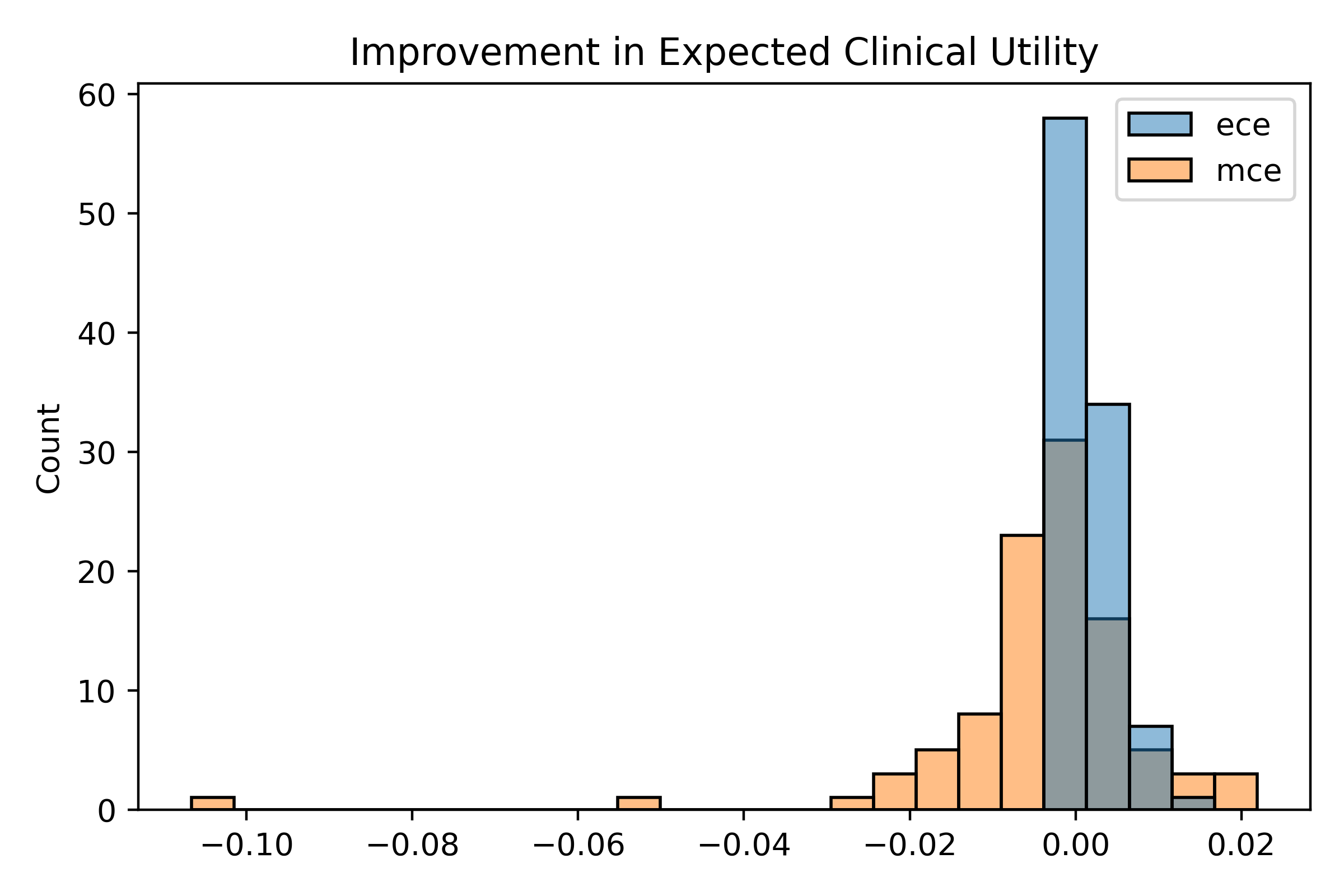}
    \caption{Comparing the gain in using the conservative therapeutic threshold w.r.t the actual calibration error (either maximum, or expected) over using the naive therapeutic threshold.}
    \label{fig:ece_mce_comparison}
\end{figure}

\textbf{Comparing  $j_{MCE}$ and $j_{ECE}$}. To evaluate the difference between computing the conservative threshold  based on the maximum or expected calibration error we conduct the following experiment: for the baseline model, we compare the gain in clinical utility (computed on the test data) from: (i) using $j_{ECE}$ w.r.t the \emph{actual} expected calibration error of the model on the test data, vs using $\js$; (ii) using $j_{MCE}$ w.r.t the \emph{actual} maximum calibration error of the model on the test data, vs using $\js$. Intuitively, since the baseline model was fairly well-calibrated on the test data, we don't expect to see a significant gain in this case. Figure \ref{fig:ece_mce_comparison} demonstrates the histogram of the gains in both cases, computed over $T=100$ independent trials (where the randomness is over the split of the data). We see that using $j_{ECE}$ leads to a gain that is indeed small (but positive), whereas using $j_{MCE}$ may often lead to a reduction in the expected clinical utility of the model. Intuitively, this happens because in the high risk regime (where there are relatively few samples under the baseline predictor) the calibration error is (naturally) larger --  and so $j_{MCE}$ takes a very worst-case perspective and ``clips'' the predictions for the low risk regime, where the model in question actually \emph{does} have clinical utility.

\textbf{Learning from small data} To demonstrate the effect of the training set size, we train models from splits of varying size. In each case, we repeat the following $T=10$ times: we partition the dataset into $S_{\text{train}}$ and  $S_{\text{test}}$, where the size of  $S_{\text{train}}$ is $s \cdot n$. We then fit the logistic model on $S_{\text{train}}$, resulting in a risk model $p$. We then compute 
 $\Delta(\alpha, s)$ as the difference between the expected Net Benefit of $p$ using $\hat{j}(\alpha, \js)$  and the expected Net Benefit of $p$ using $\js$, where the expectation is w.r.t a uniform choice of threshold $\js \sim [0,1]$. In other words, $\Delta(\alpha, s)$ measures the additive gain (in expectation) from making more conservative decisions, as reflected in $\hat{j}$. 
 
\begin{figure}[tbp]
    \centering
    \includegraphics[width=0.7\linewidth]{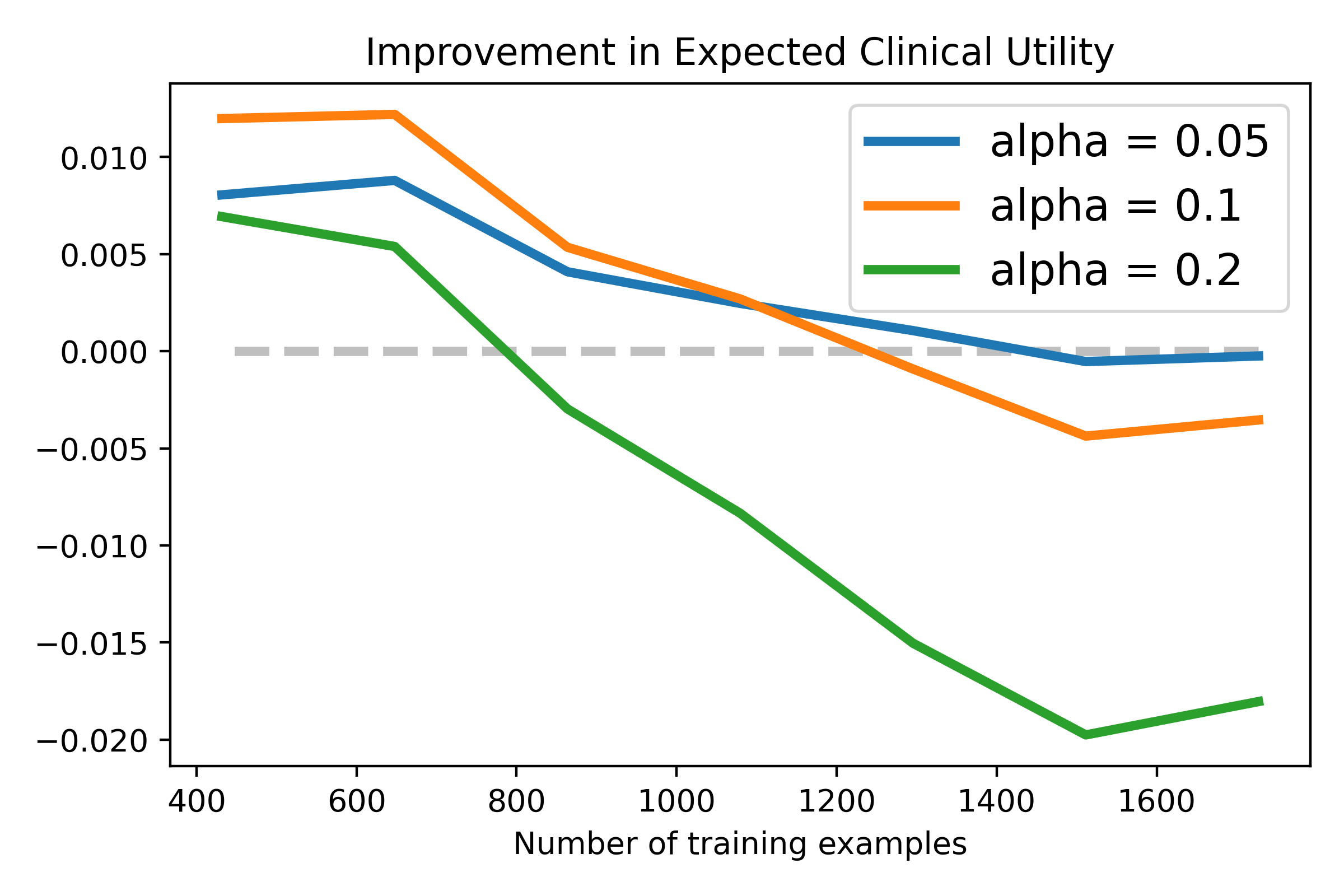}
    \caption{Plotting the difference in the Net Benefit from using the conservative therapeutic threshold and the Net Benefit from using the regular therapeutic threshold. }
    \label{fig:generalization}
\end{figure}

 Figure \ref{fig:generalization} demonstrates how $\Delta(\cdot, \cdot)$ varies as we increase the training set size $s$ from 20\% to 80\% and consider different levels of the anticipated miscalibration parameter $\alpha$: $\set{0.05, 0.1, 0.2}$. The results demonstrate that when the number of training examples is small, there is indeed clinical harm when making decisions using $\js$, which is alleviated when using $\hat{j}$. However, as the number of training examples increases, we see that using a large value of $\alpha$ is ``over conservative'' in that we end up losing a lot in clinical utility. This reflects the fact that the value of $\alpha$ should ideally represent the amount of miscalibration that we indeed anticipate we will encounter at test time.
 
\begin{figure}[tbp]
    \centering
    \includegraphics[width=1.0\linewidth]{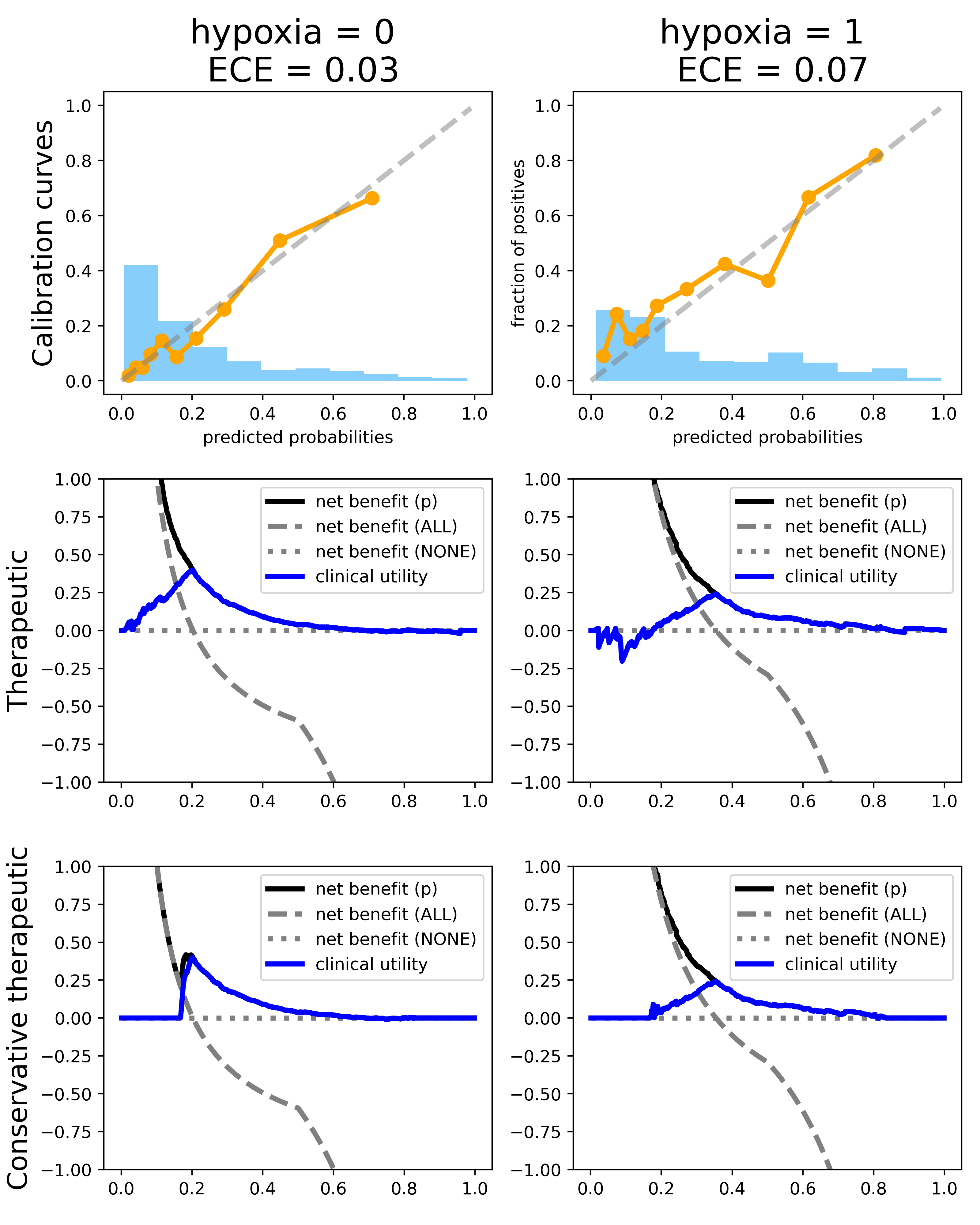}
    \caption{Evaluating a \emph{subgroup-unaware} model on the two subgroups (without hypoxia, with hypoxia) results in subgroup mis-calibration (top right) and subgroup clinical harm (middle right) when using the  therapeutic threshold $\js$. Using the conservative threshold effectively eliminates this harm. }
    \label{fig:distribution}
\end{figure}

\textbf{Distribution shift} For this experiment, we specifically consider the binary variable indicating whether an individual had hypoxia before or at admission. We refer to the subgroups as the two groups obtained by partitioning the dataset according to this feature, and we train a single \emph{subgroup-unaware} model on the entire dataset without this feature. We evaluate this model on the two subgroups, separately. 
 Figure \ref{fig:distribution} demonstrates that the model is significantly less calibrated on the group of individuals with hypoxia; as a result, it
  causes clinical harm: there are some  thresholds for which using the model with $\js$ results in a Net Benefit that is worse than having \emph{not} used the model at all. Translating predictions to decisions using the conservative threshold $j_{ECE}$ with $\alpha  = 0.2$ effectively eliminates this harm.

\section{Further Related Work}
\label{sec:related}

\textbf{Clinical decision-making.} The derivation of the therapeutic threshold dates back to \cite{pauker1975therapeutic}. Decision curve analysis was introduced by Vickers and Elkin in \cite{ vickers2006decision}, and has since gained popularity in the area of clinical risk prediction and decision making, see e.g. \cite{mallett2012interpreting, moons2012quantifying, localio2012beyond, alonzo2009clinical}. Other decision-analytic measures such as Relative Utility \cite{baker2009using} and Net Reclassification Improvement \cite{kerr2014net} are also common, but are all simple transformations of the Net Benefit \cite{ van2013evaluation}. The effect of discrimination on the Net Benefit was explored in \cite{van2013evaluation, vickers2006decision}. Recently, \cite{pfohl2022comparison, pfohl2022net} consider the impact of different fairness interventions on clinical utility.

\textbf{Calibration: clinical decision-making perspective.} Calibration is an important requirement when using risk prediction models to support medical decision making. Several studies  demonstrate that miscalibration can negatively effect a model's clinical utility. For example, Collins
and Altman \cite{collins2012predicting} evaluated prognostic models for predicting
the 10-year risk of cardiovascular disease on a large
cohort of general practice patients in the UK. Their results demonstrate that the Framingham risk score overestimates the risk in women and in men, resulting in
a harmful model for risk thresholds of around 20\%
and higher.  Additional examples in this vein are discussed in \cite{van2015calibration}, which like us, study the effect of miscalibration on Net Benefit. They use simulation studies to demonstrate that inducing miscalibration can make a model clinically harmful, stressing the importance of calibration as a desiderata for risk prediction models. Our theoretical results and experiments complement theirs: motivated by the observation that miscalibration can never be eliminated entirely,  we shift the focus to the mapping from predictions to decisions as a means to combat anticipated miscalibration. Finally, \cite{ van2016calibration} define a hierarchy of calibration measures, proving that ``moderate calibration’’, which corresponds to the standard notion of (exact) calibration, guarantees no clinical harm.  We generalize this result to the stronger notion of no regret and depart from their analysis by considering the effect of miscalibration. 

\textbf{Calibration: additional perspectives.} Calibration is an important concept that has been studied in different contexts (fairness, safety, robustness, etc). As discussed, most relevant to our work is the multi-calibration requirement proposed in 
\cite{hkrr}. Technically, it sits between the notions of  ``moderate calibration'' and ``strong calibration'' from \cite{vach2013calibration, van2016calibration}.  Multi-calibration has since been extended \cite{oi} and studied in a variety of additional contexts: ranking \cite{dwork2019rankings}, multi-group learning with general loss functions \cite{rothblum2021multi}, online decision-making \cite{gupta2021online}, and the prediction of higher order moments \cite{jung2021moment}. Most related to this work is \cite{barda2021addressing}, which applies the methodology of \cite{hkrr, kgz} to post-process risk models, such as the Framingham risk score mentioned above, to satisfy subgroup calibration w.r.t a large collection of demographic subgroups of interest. The connection between calibration and eliminating clinical harm serves as another motivation for guaranteeing predictions are also calibrated on a subgroup-level. Finally, calibration has also been studied in the context of domain generalization, see e.g. \cite{wang2020transferable, wald2021calibration}.

\textbf{Post-processing predictors.} Our focus on translating predictions to binary decisions is conceptually related to a fundamental question in learning theory, which is whether predictors that are optimal for one loss function (e.g., $\ell_2$ loss, or calibration loss) can be post-processed into predictors that are optimal for other objectives.  \cite{corbett2017algorithmic, canetti2019soft} study the power of post-processing calibrated scores into decisions when the objective is to equalize certain statistical fairness notions across subgroups, which is not our objective. More related to our work are results in  \cite{natarajan2015optimal, dembczynski2017consistency}  showing that for accuracy metrics including F-scores and AUC,
applying the optimal post-processing transformation to the $\ell_2$-loss minimizing predictor in a class $\H$ yields a classifier competitive with the best classifier in the class of binary classifiers derived by thresholding models from $\H$. Recently, \cite{gopalan2021omnipredictors} proposed the notion of an \emph{omnipredictor}, which takes this idea one step further and seeks a single predictor that can be post-processed to be competitive with a large collection of loss functions and w.r.t arbitrary classes of functions. Their work leverages connections to the notion of multicalibration to prove that this objective is computationally feasible: there exists a predictor $p$ with the guarantee that for every convex loss function, applying the optimal post-processing transformation for $\ps$ to $p$ yields an optimal classifier for this loss, albeit with a degradation in performance that depends on the Lipchitzness of the loss function in question. Our point about the clinical harm scaling like $O(\alpha\cdot m)$ under $\alpha$-miscalibration can be viewed as a lower bound showing that there indeed exist interesting and natural cases in which this is unavoidable (here, the granularity parameter $m$ acts as the Lipchitz constant). Unlike the above works, we don't assume that the distribution on which the predictor was trained is the same distribution on which performance is evaluated after post-processing. Rather, our only knowledge of the target distribution is that the predictor's miscalibration error is bounded, and we aim to minimize the worst-case regret over all target distributions and all possible alternative treatment thresholds. We further remark that the post-processing transformation we apply (the conservative threshold) is in fact \emph{not} what we would apply to $\ps$ (which has 0 miscalibration error).

\section{Discussion}
\label{sec:discuss}

Calibration is an important desideratum for risk predictor, but in practice, perfect calibration is impossible. In this work we have initiated the study of decision-making under miscalibration. Our work shows that in both theory and practice, even slight miscalibration may require taking the miscalibration into account -- and changing the method by which we ``translate'' risk predictions to binary decisions.

Our notion of the conservative therapeutic threshold is (like the therapeutic threshold) completely \emph{distribution-free}: it does not depend on the distribution or the predictor in question. This is appropriate when we want to use risk predictors that we truly know nothing about (except a guarantee on their miscalibration w.r.t the target distribution). However, there are natural scenarios in which we might envision having some kind of limited access to the target distribution (and certainly to the predictor in question). For example, we may have \emph{unlabeled} data, from which we could estimate the level sets of $p$. This suggests a different notion of regret than the one we study (since we consider the worst possible distribution), and it's interesting to explore to what extent (and how) the optimal treatment threshold under miscalibration changes when we make these additional assumptions.

\newpage
\section*{Acknowledgments} 
This project has received funding from the European Research Council (ERC) under the European Union’s Horizon 2020 research and innovation programme (grant agreement No. 819702), from the Israel Science Foundation (grant number 5219/17), from the U.S.-Israel Binational Science Foundation (grant 2018102), and from the Simons Foundation Collaboration on the Theory of Algorithmic Fairness. 
GY is also additionally supported by the Israeli Council for Higher Education (CHE) via the Weizmann Data Science Research Center, by a research grant from the Estate of Tully and Michele Plesser, and by a Google PhD fellowship. Part of this work was done when GR was visiting 
Microsoft Research and GY was at Google.

\bibliographystyle{apalike}
\bibliography{refs}

\appendix

\newpage
\clearpage
\section{Proof of Lemma \ref{lemma:net_benefit}}
\label{appendix:nb_lemma_proof}
Note that by definition, we can write the fraction of True Positives and False Positives as

\begin{align*}
    \mathtt{TP}_{\ps}(h_{p,j})=\sum_{i>j}\mu_{p,i}\cdot\ytilde, \quad
    \mathtt{FP}_{\ps}(h_{p,j})=\sum_{i>j}\mu_{p,i}\cdot(1-\ytilde)
\end{align*}

So, by definition of the symmetric Net Benefit, we have:

\begin{align*}
  \Lambda(p,j; \ps, \js)&= P'\cdot \mathtt{TP}_{\ps}(h_{p,j})-L'\cdot \mathtt{FP}_{\ps}(h_{p,j})  \\
  &= \sum_{i>j}\mu_{p,i}\left[P'\cdot\ytilde-L'\cdot(1-\ytilde)\right] \\
  &= \begin{cases}
\sum_{i>j}\mu_{p,i}\left[\frac{m-\js}{\js}\cdot\ytilde-(1-\ytilde)\right] & \js \leq m/2\\
\sum_{i>j}\mu_{p,i}\left[\ytilde-\frac{\js}{m-\js}(1-\ytilde)\right] & \js>m/2
\end{cases} \\
&= \begin{cases}
\sum_{i>j}\mu_{p,i}\frac{m\cdot \ytilde-\js}{\js} & \js\leq m/2\\
\sum_{i>j}\mu_{p,i}\frac{m\cdot \ytilde-\js}{m-\js} & \js>m/2
\end{cases} \\
&= \sum_{i>j}\mu_{p,i}\cdot\frac{m\cdot \tilde{y}_{p,\ps,i}-\js}{\min\left\{ m-\js,\,\,\js\right\} }
\end{align*}

\section{Proof of Lemma \ref{lemma:perfect}}
\label{appendix:lemma_proof}

Consider $p, \ps$ such that $(p,\ps) \in R_{ECE}(0)$. By definition, this guarantees that for every level set $i$, $\ytilde = i/m$. Using Lemma \ref{lemma:net_benefit}, we have that for every threshold $j'$,

\begin{align*}
\Lambda(p,j'; \ps, \js) - \Lambda(p,\js; \ps, \js) &= 
    \sum_{i>j'}\mu_{p,i}\cdot\frac{m\cdot \tilde{y}_{p,\ps,i}-\js}{\min\left\{ m-\js,\,\,\js\right\} } - \sum_{i>\js}\mu_{p,i}\cdot\frac{m\cdot \tilde{y}_{p,\ps,i}-\js}{\min\left\{ m-\js,\,\,\js\right\} } \\
    &= \sum_{j'<i\leq \js}\mu_{p,i}\cdot\frac{m\cdot \tilde{y}_{p,\ps,i}-\js}{\min\left\{ m-\js,\,\,\js\right\} } - \sum_{\js<i\leq j'}\mu_{p,i}\cdot\frac{m\cdot \tilde{y}_{p,\ps,i}-\js}{\min\left\{ m-\js,\,\,\js\right\} } \\
    &= \sum_{j'<i\leq \js}\underbrace{\mu_{p,i}\frac{i-\js}{\min\left\{ m-\js,\,\,\js\right\} }}_{\leq0}-\sum_{\js<i\leq j'}\underbrace{\mu_{p,i}\frac{i-\js}{\min\left\{ m-\js,\,\,\js\right\} }}_{\geq0} \\
    &\leq 0
\end{align*}

Since this holds for every threshold $j'$, it also holds for the maximum, which implies the required.

\section{Proof of Theorem \ref{thm:optimal}}
\label{appendix:thm_proof}

\subsection{Maximum Calibration Error}

\textbf{MCE: Part 1.} Recall that we defined $c(j; \js, R) \triangleq \max_{\set{\mu_{p,i}}_{i=1}^m \in \Xi, \,\, (p,p^{\star})\in R(\alpha)} \mathtt{regret}(p,j; \ps, \js)$. In this section we will use $\tilde{c}(j;\js, R)$ to denote the maximal regret when $p$ is additionally constrained to be a constant predictor. Our objective in this part is to show that for every $j$ and $\js$, $c(j;\js, R_{\textbf{MCE}})=\tilde{c}(j;\js, R_{\textbf{MCE}})$. Note that the direction $\tilde{c}(j;\js, R_{\textbf{MCE}}) \leq c(j;\js, R_{\textbf{MCE}})$ is trivial; it's left to prove the other direction.

Fix $j,\js$ and $\alpha > 0$. We want to show that for every $\set{\mu_{p,i}}_{i=1}^m \in \Xi$ and predictors $p, \ps$  such that $(p,\ps)\in R_{MCE}(\alpha)$, and for every  $j_{R}$, there exist $\set{\mu_{\tilde{p},i}}_{i=1}^m \in \Xi$ and predictors $\tilde{p}, \tilde{\ps}$ and $\tilde{j}_{R}$ such that:

\begin{align}
    \label{eqn:simple_p_0}  (\tilde{p},\tilde{p}^{*}) &\in R_{MCE}(\alpha) \\
  \label{eqn:simple_p_1}  \exists i \text{\,\,\, such that\,\,\,} \mu_{\tilde{p},i} &= 1 \\
  \label{eqn:simple_p_2}  \mathtt{regret}(j;\js,p,\ps,j_{R}) &\leq \mathtt{regret}(j;\js,\tilde{p},\tilde{\ps},\tilde{j}_{R})
\end{align}

Fix $\set{\mu_{p,i}}_{i=1}^m \in \Xi$ and predictors $p, \ps$  such that $(p,\ps)\in R_{MCE}(\alpha)$, and fix a threshold $j_R$. W.l.o.g, assume $j_{R}<j$. Let $i^{*}=\arg\max_{j_{R}<i<j}\frac{m\cdot\tilde{y}_{p,p^{*},i}-j^{*}}{\min\left\{ m-j^{*},j^{*}\right\} }$, and define the following predictors:

\begin{align*}
    \tilde{p}(x) &\equiv i^{*}/m \\
    \tilde{p}^{*}(x) &\equiv\tilde{y}_{p,p^{*},i^{*}}
\end{align*}

Note that $\tilde{p}$ is a constant prediction, so (\ref{eqn:simple_p_1}) holds. Also, (\ref{eqn:simple_p_0}) follows directly by the assumption that $(p,\ps)\in R_{MCE}(\alpha)$ (but note that here we use the fact that miscalibration is measured using \emph{maximum} calibration error). We will now argue that taking $\tilde{j}_{R}=j_{R}$, also (\ref{eqn:simple_p_2}) holds. Note that by definition of $i^{\star}$,

\begin{align*}
    \mathtt{regret}(j;\js,p,\ps,j_{R})	&=\sum_{i>j_{R}}\mu_{p,i}\cdot\frac{m\cdot\tilde{y}_{p,\ps,i}-\js}{\min\left\{ m-\js,\js\right\} }-\sum_{i>j}\mu_{p,i}\cdot\frac{m\cdot\tilde{y}_{p,\ps,i}-\js}{\min\left\{ m-\js,\js\right\} } \\
	&\leq \sum_{j_{R}<i<j}\mu_{p,i}\cdot\frac{m\cdot\tilde{y}_{p,\ps,i^{*}}-\js}{\min\left\{ m-\js,\js\right\} } \\
	&=\frac{m\cdot\tilde{y}_{p,\ps,i^{*}}-\js}{\min\left\{ m-\js,\js\right\} }\cdot\underbrace{\sum_{j_{R}<i<j}\mu_{p,i}}_{\leq1} \\
	&\leq\frac{m\cdot\tilde{y}_{p,\ps,i^{*}}-\js}{\min\left\{ m-\js,\js\right\} }
\end{align*}

On the other hand, by the definition of $\tilde{p}$ and $\tilde{\ps}$, we have:

\begin{align*}
    \mathtt{regret}(j;\js,\tilde{p},\tilde{p}^{*},\tilde{j}_{R})	&=\sum_{i>j_{R}}\mu_{\tilde{p},i}\cdot\frac{m\cdot\tilde{y}_{\tilde{p},\tilde{p}^{*},i}-j^{*}}{\min\left\{ m-\js,\js\right\}  }-\sum_{i>j}\mu_{\tilde{p},i}\cdot\frac{m\cdot\tilde{y}_{\tilde{p},\tilde{p}^{*},i}-\js}{\min\left\{ m-\js,\js\right\} } \\
	&=\sum_{j_{R}<i<j}\mu_{\tilde{p},i}\cdot\frac{m\cdot\tilde{y}_{\tilde{p},\tilde{p}^{*},i}-\js}{\min\left\{ m-\js,\js\right\} } \\
	&=\frac{m\cdot\tilde{y}_{\tilde{p},\tilde{p}^{*},i^{*}}-\js}{\min\left\{ m-\js,\js\right\} } \\
	&=\frac{m\cdot \E_{x}[\tilde{p}^{*}(x)]-\js}{\min\left\{ m-\js,\js\right\} } \\
	&=\frac{m\cdot\tilde{y}_{p,p^{*},i^{*}}-\js}{\min\left\{ m-\js,\js\right\} }
\end{align*}
Combining, we have shown  (\ref{eqn:simple_p_2}), which concludes the claim that $ c(j;\js, R_{\textbf{MCE}}) \leq \tilde{c}(j;\js, R_{\textbf{MCE}}) $.

\textbf{MCE: Part 2}. Next, we will show that the maximal regret w.r.t $(p, \ps) \in R_{MCE}(\alpha)$ when $p$ is constant is

\begin{equation}
\label{eqn:cost_mce}
    \frac{\max\left\{ \min\left\{ m-\js,j-\js+\alpha m\right\} ,\quad\min\left\{ \js,\js-j+\alpha m\right\} \right\} }{\min\{m-\js,\js\}}
\end{equation}

Fix a constant predictor $p$ and an arbitrary predictor $\ps$. Let $v$ denote the value $p$ is supported on (i.e., the level set for which that $\mu_{p,v} = 1$). 
Note that w.l.o.g, we can additionally assume that $\ps$ is constant too - the regret consists of the difference between Net Benefits, which is only a function of $\tilde{y}_{p,\ps, v}$, which is the same if we replace $\ps$ with its expectation. We use $v^\star$ to denote the value $\ps$ is supported on. In this case, the regret of $j$ w.r.t $\js$ is exactly

\begin{equation*}
    \begin{cases}
0 & v<j,\vs<\js \\
\card{\frac{m\cdot \vs-\js}{\min\{m-\js,\js\}}} & v<j,\vs\geq \js\\
\card{\frac{m\cdot \vs-\js}{\min\{m-\js,\js\}}} & v\geq j,\vs<\js\\
0 & v\geq j,\vs \geq \js
\end{cases}
\end{equation*}

Recall that to determine the cost of $j$, we would like to choose $v, \vs$ to maximize this expression. We will consider two choices for $v$: $v \leq j$ and $v > j$:

\begin{enumerate}
    \item When $v \leq j$, to maximize the regret we want to choose $\vs \geq j$. In that case the regret will be $\card{\frac{m\cdot \vs-\js}{\min\{m-\js,\js\}}}$, which is maximized when $\vs$ is maximized; under the maximum calibration constraint, $m\cdot \vs$ can be as large as $\min \set{m, v + \alpha m}$, which in turn can be as large as $\min \set{m, j + \alpha m}$. So the regret in this case is
    \begin{equation*}
        \frac{\min\{m,j+\alpha m\}-\js}{\min\{m-\js,\js\}}=\frac{\min\{m-\js,j-\js+\alpha m\}}{\min\{m-\js,\js\}}
    \end{equation*}
    
     \item When $v > j$, to maximize the regret we want to choose $\vs < j$. In that case the regret will be $\card{\frac{\js-m\cdot \vs}{\min\{m-\js,\js\}}}$, which is maximized when $\vs$ is minimized; under the maximum calibration constraint, $m\cdot \vs$ can be as small as $\max\{0,v-\alpha m\}$, which in turn can be as small as $\max\{0,j-\alpha m\}$. So the regret in this case is
     
     \begin{equation*}
         \frac{\max\{0,j-\alpha m\}-\js}{\min\{m-\js,\js\}}=\frac{\max\{-\js,j-\js-\alpha m\}}{\min\{m-\js,\js\}}=\frac{\min\{\js,\js-j+\alpha m\}}{\min\{m-\js,\js\}}
     \end{equation*}
\end{enumerate}

Finally, since we can also chose whether $v\leq j $ or $v > j$, we obtain that the overall maximum regret is exactly the expression in Equation (\ref{eqn:cost_mce}), which concludes this part.

\textbf{MCE: Part 3}. It's left to argue why the minimizer of the cost from Equation (\ref{eqn:cost_mce}) is $\hat{j}$ defined in the theorem statement. First, we note that since the denominator of Equation (\ref{eqn:cost_mce}) is non-negative and independent of $j$, it suffice to minimize the following cost

\begin{equation*}
    \max\left\{ \min\left\{ m-\js,j-\js+\alpha m\right\} ,\quad\min\left\{ \js,\js-j+\alpha m\right\} \right\} 
\end{equation*}

Note:

\begin{itemize}
    \item In the first minimum, the first expression ``dominates'' (i.e., is smallest) iff $j\geq(1-\alpha)m$;
    
    \item In the second minimum, the first expression ``dominates'' (i.e., is smallest) iff $j\leq\alpha m$.
\end{itemize}

So we can start by simplifying according to the following regions:
Left (meaning $j \leq \alpha \cdot m$), Middle (meaning $\alpha m\leq j\leq(1-\alpha)m$) and Right (meaning $j \geq (1-\alpha)m$). 
\begin{itemize}

    \item \textbf{Middle}: The expression becomes $\max\left\{ j-\js+\alpha m,\quad \js-j+\alpha m\right\}$. The first term increases in $j$ and the second decreases in $j$, so the optimal choice is $j-\js+\alpha m=\js-j+\alpha m\Rightarrow j=\js$. 
    \item  \textbf{Left}:  The expression becomes $\max\left\{ j-\js+\alpha m,\quad \js\right\}$. Similarly, the optimal choice in this range is $j=2\js-\alpha m$.
    \item \textbf{Right}: The expression becomes $\max\left\{ m-\js,\quad \js-j+\alpha m\right\}$,  so the optimal choice is $j=2\js-(1-\alpha)m$.
\end{itemize}

To summarize, given $\js$, we have several options: choose $j=\js$ incurring a cost of $\alpha m$; choose $j=2\js-\alpha m$ incurring a cost of $\js$; or choose $j=2\js-(1-\alpha)m$ incurring a cost of $m-\js$. From this, we can derive the optimal threshold $\hat{j}$, depending on $\js$: 

\begin{itemize}
    \item if $\alpha m\leq \js\leq(1-\alpha)m$, we have that $\js\geq\alpha m$ and $m-\js\geq\alpha m$, so the optimal choice is $j=\js$.
    \item if $\js\leq\alpha m$, then the optimal choice is $j=\max\{0,\,\,2\js-\alpha m\}$.
    \item If $\js\geq(1-\alpha)m$, then the optimal choice is $j=\min\{1,\,\,2\js-(1-\alpha)m\}$.
\end{itemize}

Which is precisely the expression in the theorem statement, concluding the proof of the optimal threshold under miscalibration when using maximum calibration error.

\subsection{Expected Calibration Error}

We now follow the same logic but when miscalibration is measured using \emph{expected} calibration error (ECE). Here, the reduction in the first part is more subtle: we show that we also need to take into account slightly more complex predictors, that can be supported on two values (not one).

\textbf{ECE: Part 1}. 
Fix $j,\js$ and $\alpha > 0$.  This time, we want to show that for every $\set{\mu_{p,i}}_{i=1}^m \in \Xi$ and predictors $p, \ps$  such that $(p,\ps)\in R_{ECE}(\alpha)$, and for every  $j_{R}$, there exist $\set{\mu_{\tilde{p},i}}_{i=1}^m \in \Xi$ and predictors $\tilde{p}, \tilde{\ps}$ and $\tilde{j}_{R}$ such that:

\begin{align}
    \label{eqn:less_simple_p_0}  (\tilde{p},\tilde{p}^{*}) &\in R_{ECE}(\alpha) \\
  \label{eqn:less_simple_p_1}  \exists i_1,i_2 \text{\,\,\, such that\,\,\,} \mu_{\tilde{p},i_1} + \mu_{\tilde{p},i_2} &= 1 \\
  \label{eqn:less_simple_p_2}  \mathtt{regret}(j;\js,p,\ps,j_{R}) &\leq \mathtt{regret}(j;\js,\tilde{p},\tilde{\ps},\tilde{j}_{R})
\end{align}

Fix $\set{\mu_{p,i}}_{i=1}^m \in \Xi$ and predictors $p, \ps$  such that $(p,\ps)\in R_{ECE}(\alpha)$, and fix a threshold $j_R$.
 W.l.o.g, assume $j_{R}<j$. Intuitively, our construction of the ``simple'' $\tilde{p}$ will collect all the level sets outside $[j_R, j]$ into a single level set, whose value is arbitrary (say 0), and all the level sets inside $[j_R, j]$ into a single level set, whose value is the mean. We will construct $\tilde{\ps}$ accordingly to ensure that the resulting pair $(\tilde{p}, \tilde{\ps})$ is still in the ECE relation. Consider the predictor $\tilde{p}$ that puts $\mu_1$ mass on a value $v_1$ and $\mu_2$ mass on a value $v_2$, defined as follows:

\begin{align*}
    v_1 &= 0, &\quad  \mu_1 = \sum_{i \notin [j_R, j]} \mu_{p,i} \\
    v_2 &= \frac{\sum_{i \in [j_R, j]} \mu_{p,i}\cdot(i/m)}{\sum_{i \in [j_R, j]} \mu_{p,i}},   &\quad \mu_2 = \sum_{i \in [j_R, j]} \mu_{p,i}
\end{align*}

For $\tilde{\ps}$, we define it such that $\tilde{y}_{\tilde{p}, \tilde{\ps}, v_1} = \vs_1$ and $\tilde{y}_{\tilde{p}, \tilde{\ps}, v_2} = \vs_2$, where $\vs_1, \vs_2$ are defined as follows:

\begin{align*}
    \vs_1 &= 0 \\
    \vs_2 &= \frac{\sum_{i \in [j_R, j]} \mu_{p,i}\cdot \ytilde}{\sum_{i \in [j_R, j]} \mu_{p,i}}
\end{align*}

Note that Equation (\ref{eqn:less_simple_p_1}) holds since by construction $\mu_1 + \mu_2 = 1$. For (\ref{eqn:less_simple_p_0}), we bound the ECE of $\tilde{p}$ w.r.t $\tilde{\ps}$:

\begin{align*}
    \sum_i \mu_{\tilde{p}, i} \cdot \card{ i/m - \tilde{y}_{\tilde{p}, \tilde{\ps}, i}} &= \mu_1 \cdot \card{v_1 - \vs_1} +  \mu_2 \cdot \card{v_2 - \vs_2} \\
    &= \mu_2 \cdot \card{v_2 - \vs_2} \\ 
    &= \card{\sum_{i \in [j_R, j]} \mu_{p,i}\cdot(i/m) - \sum_{i \in [j_R, j]} \mu_{p,i}\cdot \ytilde} \\
    &\leq \sum_{i \in [j_R, j]} \mu_{p,i}\cdot \card{i/m - \ytilde} \\
    &\leq \alpha
\end{align*}

where the last transition follows by the fact that $(p,\ps)\in R_{ECE}(\alpha)$. Finally, for Equation (\ref{eqn:less_simple_p_2}):

\begin{align*}
    \mathtt{regret}(j;\js,\tilde{p},\tilde{\ps},\tilde{j}_{R})	
	&=\sum_{j_{R}<i<j}\mu_{\tilde{p},i}\cdot\frac{m\cdot\tilde{y}_{\tilde{p},\tilde{\ps},i}-\js}{\min\left\{ m-\js,\js\right\} } \\ 
	&= \mu_2 \cdot \frac{m\cdot\vs_2-\js}{\min\left\{ m-\js,\js\right\} } \\
	&= \sum_{i \in [j_R, j]} \mu_{p,i} \cdot \frac{m\cdot\frac{\sum_{i \in [j_R, j]} \mu_{p,i}\cdot \ytilde}{\sum_{i \in [j_R, j]} \mu_{p,i}}-\js}{\min\left\{ m-\js,\js\right\} } \\
	&= \sum_{i \in [j_R, j]} \mu_{p,i} \cdot \frac{m\cdot\frac{\sum_{i \in [j_R, j]} \mu_{p,i}\cdot \ytilde}{\sum_{i \in [j_R, j]} \mu_{p,i}}-\frac{\js \cdot \sum_{i \in [j_R, j]}\cdot \mu_{p,i} }{\sum_{i \in [j_R, j]}\cdot \mu_{p,i}}}{\min\left\{ m-j^{*},j^{*}\right\} } \\
	&= \frac{ \sum_{i \in [j_R, j]} \mu_{p,i}\cdot \br{ m \cdot \ytilde - \js}}{\min\left\{ m-\js,\js\right\} } \\
	&= \mathtt{regret}(j;\js,p,\ps,j_{R}) 
\end{align*}

This concludes the proof of Part 1.

\textbf{ECE: Part 2}.  Next, we will show that the maximal regret incurred by a threshold $j$ w.r.t $\js$, under $(p, \ps) \in R_{ECE}(\alpha)$ when $p$ is additionally constrained to be ``almost constant'' (in the sense formalized above) is

\begin{equation}
\label{eqn:max_regret_almost_constant} \frac{1}{\min \set{m-\js, \js}} \cdot
    \begin{cases}
\max\left\{ \js,\quad\alpha m\cdot\frac{m-\js}{m-j}\right\}  & j<\js,\quad j\leq\alpha m\\
\max\left\{ \js-j+\alpha m,\quad\alpha m\cdot\frac{m-\js}{m-j}\right\}  & \alpha m<j\leq\js\\
\max\left\{ j-\js+\alpha m,\quad\alpha m\cdot\frac{\js}{j}\right\}  & \js<j\leq(1-\alpha)m\\
\max\left\{ m-\js,\quad\alpha m\cdot\frac{\js}{j}\right\}  & j\geq\js,\quad j>(1-\alpha)m
\end{cases}
\end{equation}

Fix $j, \js$. Consider some choice of threshold $j_R$ and predictors $p, \ps \in R_{ECE}(\alpha)$. Since $p$ is assumed to be ``simple'', we will assume it is supported on two values $v_1, v_2$ (where $v_1$ is between $j$ and $j_R$, and $v_2$ is not), with $\vs_1, \vs_2$ denoting the respective conditional expectation of $\ps$, similar to the notation used in Part 1. We will split into two cases: $j_R \leq j$ and $j_R > j$.

When $j_R \leq j$, 

\begin{equation*}
    \min \set{m-\js, \js} \cdot  \mathtt{regret}(j;\js,p,\ps,j_{R}) = \sum_{j_{R}<i<j}\mu_{p,i}\cdot(m\tilde{y}_{p,\ps,i}-\js) = \mu_{1}\cdot(m\cdot \vs_1-\js)
\end{equation*}

Since we want to maximize this expression, we want to choose $\vs_1$ as large as possible; we can therefore assume w.l.o.g that $\vs_1 \geq v_1$ (and also trivially $\vs_2 \geq v_2$). We can therefore re-parametrize the expressions in terms of $v_1, v_2$ and $\delta_1 = \vs_1 - v_1 > 0$, $\delta_2 = \vs_2 - v_2 > 0$, yielding the following optimization problem (whose value is the maximal regret we are interested in):

\begin{equation*}
    \max_{\mu_{1},v_{1},v_{2},\delta_{1},\delta_{2}}\mu_{1}\cdot(m\cdot(v_{1}+\delta_{1})-\js)\text{ subject to \ensuremath{\begin{cases}
v_{1} & \in[j_{R},j]\\
v_{2} & \notin[j_{R},j]\\
\mu_{1} & \in[0,1]\\
\delta_{1} & \in[0,1-v_{1}]\\
\delta_{2} & \in[0,1-v_{2}]\\
\mu_{1}\cdot\delta_{1}+(1-\mu_{1})\cdot\delta_{2} & \leq\alpha
\end{cases}}}
\end{equation*}

Note that we can re-write the objective as $m\cdot\mu_{1}v_{1}+m\cdot\mu_{1}\delta_{1}-\mu_{1}\cdot \js$. Since $\mu_{1}\delta_{1}$ will just be $\alpha$ (or as large as it can be), we are left with maximizing $m\cdot\mu_{1}(v_{1}-\js/m)$. Now, the behaviour will depend on the relationship between $j$ and $\js$:

\begin{itemize}
    \item If $j\geq \js$, then the optimal thing is to put the most weight on $\mu_{1}$ as possible; i.e., have $v_{1}=j/m$, $\mu_{1}=1$ and $\delta_{1}=\alpha$. For the solution to be legal, $v_{1}+\delta_{1}$ must not exceed $1$. So the optimal choice is $\delta_{1}=\min\left\{ \alpha,1-j/m\right\}$ . In particular:
    
    \begin{itemize}
        \item If $\alpha<1-j/m$, the value of the objective would be $j-\js+\alpha m$.
        
        \item If $\alpha>1-j/m$, the value of the objective would be $m-\js$.
    \end{itemize}
    
    \item If $j<\js$, then the optimal thing would be to put as little weight on $\mu_{1}$ as possible; i.e. have $v_{1}=j$, $\delta_{1}=1-j/m$ and $\mu_{1}=\frac{\alpha}{1-j/m}$. The value of the objective would be $\frac{\alpha}{1-j/m}\cdot(m(j/m+1-j/m)-\js)=\alpha m\cdot\frac{m-\js}{m-j}$.
\end{itemize}

To summarize, the maximal regret achievable when $j_R \leq j$ is

\begin{equation}
\label{eqn:j_R_leq_j}
    \frac{1}{\min \set{m-\js, \js}} \cdot =\begin{cases}
j-\js+\alpha m & \js<j<(1-\alpha)m\\
m-\js & j>\js,\quad j>(1-\alpha)m\\
\alpha m\cdot\frac{m-\js}{m-j} & j<\js
\end{cases}
\end{equation}

Let us now repeat this analysis for the case $j_R > j$. This time,

\begin{equation*}
    \min \set{m-\js, \js} \cdot  \mathtt{regret}(j;\js,p,\ps,j_{R}) = \sum_{j<i<j_R}\mu_{p,i}\cdot(m\tilde{y}_{p,\ps,i}-\js) = \mu_{1}\cdot(\js - m\cdot \vs_1)
\end{equation*}

Since we want to maximize this expression, this time we want to choose $\vs_1$ as small as possible; we can therefore assume w.l.o.g that $\vs_1 \leq v_1$ (and also trivially $\vs_2 \leq v_2$). Re-parametrizing in terms of $\delta_1 = v_1 - \vs_1 \geq 0$ and $\delta_2 = v_2 - \vs_2 \geq 0$, we have the following optimization problem:

\begin{equation*}
    \max_{\mu_{1},v_{1},v_{2},\delta_{1},\delta_{2}}\mu_{1}\cdot(\js-m\cdot(v_{1}-\delta_{1}))\text{ subject to \ensuremath{\begin{cases}
v_{1} & \in[j,j_{R}]\\
v_{2} & \notin[j,j_{R}]\\
\mu_{1} & \in[0,1]\\
\delta_{1} & \in[0,v_{1}]\\
\delta_{2} & \in[0,v_{2}]\\
\mu_{1}\cdot\delta_{1}+(1-\mu_{1})\cdot\delta_{2} & \leq\alpha
\end{cases}}}
\end{equation*}

Since we can re-write the objective as $\mu_{1}\cdot \js+m\cdot\mu_{1}\delta_{1}-m\cdot\mu_{1}v_{1}$, and again $\mu_{1}\delta_{1}$ will just be $\alpha$ (or as large as it can be), we are left with maximizing $m\cdot\mu_{1}(\js/m-v_{1})$. The behaviour again depends on the relationship between $j$ and $\js$:

\begin{itemize}
    \item If $j<\js$, then the optimal thing is to put the most weight on $\mu_{1}$ as possible; i.e., have $v_{1}=j/m$, $\mu_{1}=1$ and $\delta_{1}=\alpha$. For the solution to be legal, $v_{1}-\delta_{1}$ must be non-negative. So the optimal feasible choice is  $\delta_{1}=\min\left\{ \alpha,j/m\right\}$. In particular: 
    \begin{itemize}
        \item If $\alpha<j/m$, the value of the objective will be $\js-j+\alpha m$.
        \item If $\alpha>j/m$, the value of the objective will be $\js$.
        
    \end{itemize}
    
    \item If $j>\js$, then the optimal thing is to put as little weight on $\mu_{1}$ as possible; i.e. have $v_{1}=j/m$, $\delta_{1}=j/m$ and $\mu_{1}=\frac{\alpha}{j/m}$. The value of the objective would be $\frac{\alpha}{j/m}\cdot(\js-m(j/m-j/m))=\alpha m\cdot\frac{\js}{j}$.
\end{itemize}

To summarize, the maximal regret achievable when $j_R > j$ is

\begin{equation}
\label{eqn:j_R_geq_j}
    \frac{1}{\min \set{m-\js, \js}} \cdot =\begin{cases}
\js-j+\alpha m & \alpha m<j<\js\\
\js & j<\js,\quad j<\alpha m\\
\alpha m\cdot\frac{\js}{j} & j>\js
\end{cases}
\end{equation}

Finally, since we also maximize over $j_R$, we combine Equations (\ref{eqn:j_R_leq_j}) and (\ref{eqn:j_R_geq_j}) to obtain precisely the expression from Equation (\ref{eqn:max_regret_almost_constant}), which concludes the proof of this part.

\textbf{ECE: Part 3}. It's left to argue that the minimizer of the cost in Equation (\ref{eqn:max_regret_almost_constant}) is the expression from the theorem statement. We will ignore the term $\frac{1}{\min \set{m-\js, \js}}$ since it is non-negative and does not depend on $j$. 

As we did for MCE, we will begin by simplifying the cost according to the following regions: Left (meaning $j \leq \alpha\cdot m$), Left Middle (meaning $\alpha m<j\leq\js$), Right Middle (meaning $\js\leq j<(1-\alpha)m$), and Right (meaning $j>(1-\alpha)m$).

\begin{itemize}
    \item \textbf{Left}: The expression becomes $\max\{\js,\alpha m\cdot\frac{m-\js}{m-j}\}$, and the optimal choice in this region is $\max\{0,\,\,(1+\alpha)m-\frac{\alpha m^{2}}{\js}\}$.
    \item \textbf{Left Middle}: The expression becomes $\max\{\js-j+\alpha m,\,\,\alpha m\cdot\frac{m-\js}{m-j}\}$, and $\js$ is an optimal choice in this region.\footnote{This follows by observing that $\js$ is a solution to the quadratic equation (in $j$) $\js-j+\alpha m=\alpha m\cdot\frac{m-\js}{m-j}$}
    \item \textbf{Right Middle}:  The expression becomes $\max\{j-\js+\alpha m,\,\,\alpha m\cdot\frac{\js}{j}\}$, and $\js$ is an optimal choice in this region.\footnote{This follows by observing that $\js$ is a solution to the quadratic equation (in $j$) $j-\js+\alpha m=\alpha m\cdot\frac{\js}{j}$}
    \item \textbf{Right}:The expression becomes $\max\{m-\js,\alpha m\cdot\frac{\js}{j}\}$, and the optimal choice in this region is $\min\{m,\,\,\alpha m\cdot\frac{\js}{m-\js}\}$.
\end{itemize}

To summarize, given $\js$, we have several options: choose $\max\{0,\,\,(1+\alpha)m-\frac{\alpha m^{2}}{\js}\}$, incurring a cost of $\js$; choose $\js$, incurring a cost of $\alpha m$; or choose $\min\{m,\,\,\alpha m\cdot\frac{\js}{m-\js}\}$, incurring a cost of $m-\js$. From this, we can derive the optimal threshold $\hat{j}$ as a function of $\js$, which exactly follows that in the theorem statement, concluding the required.

\end{document}